\documentclass{article}
\usepackage{amssymb}
\usepackage{graphicx,epsfig,setspace,subfig,url,amsmath}
\usepackage{algorithm}
\usepackage{algorithmic}
\usepackage{longtable}
\usepackage{array}
\usepackage{amsmath}
\usepackage{mathtools}
\usepackage{multirow}
\usepackage{relsize}
\usepackage{lineno}
\usepackage[final]{nips_2017}

\usepackage[T1]{fontenc}
\usepackage[font=small,labelfont=bf,tableposition=top]{caption}

\urldef{\mailsb}\path|{gustavo.carneiro@adelaide.edu.au}|

 \normalsize

\let\svthefootnote\thefootnote
\newcommand\freefootnote[1]{%
  \let\thefootnote\relax%
  \footnotetext{#1}%
  \let\thefootnote\svthefootnote%
}

\title{Producing radiologist-quality reports for interpretable artificial intelligence.}

\author{
  William Gale\footnotemark, Gustavo Carneiro \\
  School of Computer Science \\
  The University of Adelaide\\
  Adelaide, SA 5000 \\
  \texttt{will@wgale.com} \\
  \texttt{gustavo.carneiro@adelaide.edu.au}\\
  \And
  Luke Oakden-Rayner\footnotemark[\value{footnote}], Lyle J. Palmer \\
  School of Public Health\\
  The University of Adelaide\\
  Adelaide, SA 5000 \\
  \texttt{\{luke.oakden-rayner,lyle.palmer\}} \\
  \texttt{@adelaide.edu.au}\\
  \And
  Andrew P. Bradley \\
  Faculty of Science and Engineering \\
  Queensland University of Technology \\
  Brisbane, QLD 4001 \\
  \texttt{a6.bradley@qut.edu.au} \\
}

\begin{document}
\freefootnote{* These authors contributed equally to the work}

\maketitle


\begin{abstract}

Current approaches to explaining the decisions of deep learning systems for medical tasks have focused on visualising the elements that have contributed to each decision.
We argue that such approaches are not enough to ``open the black box'' of medical decision making systems because they are missing a key component that has been used as a standard communication tool between doctors for centuries: language.
We propose a model-agnostic interpretability method that involves training a simple recurrent neural network model to produce descriptive sentences to clarify the decision of deep learning classifiers. 
We test our method on the task of detecting hip fractures from frontal pelvic x-rays. This process requires minimal additional labelling despite producing text containing elements that the original deep learning classification model was not specifically trained to detect.
The experimental results show that: 1) the sentences produced by our method consistently contain the desired information, 2) the generated sentences are preferred by doctors compared to current tools that create saliency maps, and 3) the combination of visualisations and generated text is better than either alone.

\end{abstract}

\section{Introduction}
\label{sec:intro}

Recent advances in machine learning techniques have resulted in medical decision making systems that equal human doctors at a variety of tasks~\cite{gulshan2016development,esteva2017dermatologist}, but significant barriers to clinical implementation remain. One widely discussed issue in the machine learning community, has been the concept of interpretability~\cite{lipton2016mythos} -- how believable and understandable the decisions of machine learning models are. Indeed, 
the European Union has recently passed legislation requiring machine learning systems to be explainable if their decisions affect humans~\cite{goodman2016european}.
In response to these concerns, researchers have sought to ``open the black box of AI''~\cite{castelvecchi2016can} by attempting to explain the decisions of systems in a variety of ways, which typically have fallen into the following categories: 1) identifying regions of the image that contribute to the decision (saliency mapping/heatmaps)~\cite{smilkov2017smoothgrad}, 2) visualising the learned features of deep neural networks~\cite{mordvintsev2015inceptionism}, and 3) identifying clusters of samples that receive the same decisions, for human exploration~\cite{maaten2008visualizing}.

While these methods provide some insights into the inner workings of deep neural networks, they have yet to convincingly address concerns around model interpretability. We believe that this is because there are two competing broad definitions of interpretability. Machine learning specialists seek to understand the mechanisms underlying their models, but doctors and other end-users simply want ``human-style'' explanations; the same sort of explanations they currently receive from other humans. 
Humans explain their decisions with natural language descriptions, either in the form of speech or a written report. While this description is often post-hoc and may not accurately reflect the decision making process, it is the form of explanation that end-users currently expect and are satisfied with. For example, if a human radiologist says ``there is an irregular mass with abnormal blood flow that is consistent with cancer'' this description both communicates the relevant finding, but also describes the important image features that informed that decision. The clinician receiving the report can judge whether the description matches the decision, and whether the description matches their own interpretation of the image.

There have been multiple previous attempts to produce descriptive text from natural images~\cite{shin2016deep,xu2015show} and radiology reports~\cite{krening2017learning,wang2018tienet}, but these methods were employed to make use of readily available free-text data as a training signal with the goal of improving model performance. In the case of the radiology tasks, attempting to reproduce highly variable radiology reports resulted in outputs that appear too inconsistent to be useful as explanations. Even in the most cutting edge work in the field~\cite{wang2018tienet} the text produced is far from human-like and would not satisfy doctors as either a diagnostic report, or as an explanation for model decisions.

In this paper, we are motivated by the idea that the explanation given by a doctor for any narrow medical task is often fairly simple, and the variability in the phrasing of these reports does not change their meaning. Instead of trying to reproduce whole reports, or even just the relevant report sentences, we attempt to train an image-to-text model to produce simplified but meaningful descriptions that describe and justify a diagnostic decision in a way that is satisfying to clinicians.

In this work we present a model-agnostic extension of deep learning classifiers based on a recurrent neural network model with a visual attention mechanism. This model produces a short text description to explain the medical decisions made by the classifier.
We test our tool on a large dataset of hip fractures in pelvic x-rays, and show that our model can produce high quality ``human-style'' explanations. 

\vspace{-.1in}  
\section{Materials and Methods}
\vspace{-.1in}  
\label{sec:materials_methods}

\subsection{Materials}
\label{sec:materials}

\begin{figure}
\begin{center}
\begin{tabular}{c}
\includegraphics[width=0.95\textwidth]{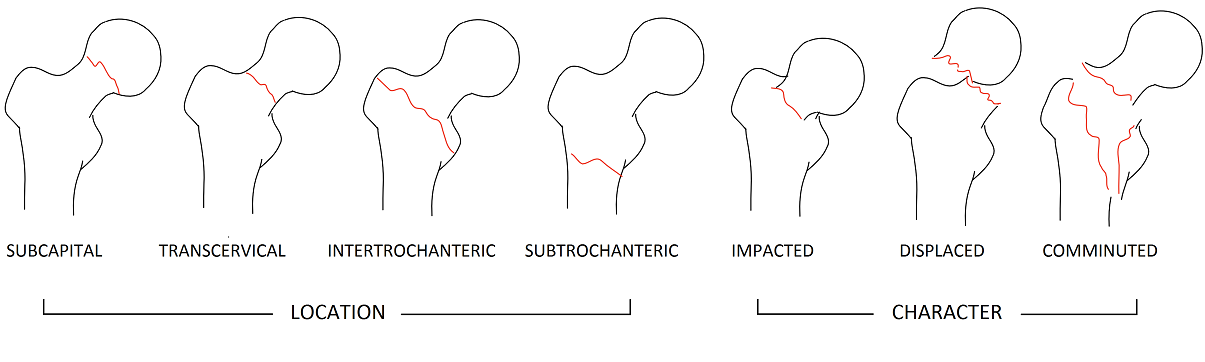} \\
\end{tabular}
\end{center}
\vspace{-.2in} 
\caption{Key terms describing the location and character of hip fractures.}
\label{fig:fractures1}
\end{figure}

The dataset available for this work consists of 50,363 frontal pelvic X-rays, containing 4,010 hip fractures~\cite{gale2017detecting}.  These images were randomly divided into a training set (41,032 images, with 2923 hip fractures), a validation set (4,754 images, with 414 hip fractures) for model selection, and a held-out test set (4,577 images, with 348 hip fractures). There was no overlap of patients between the three sets.
The deep learning classifier explored in our paper is the recently developed DenseNet~\cite{huang2017densely}, trained to classify hip fractures from frontal pelvic X-rays~\cite{gale2017detecting}. This model has been shown to produce a classification performance equivalent to a human radiologist with an area under the ROC curve of 0.994~\cite{gale2017detecting}.

The original training dataset for the CNN consisted of images with labels which included the anatomical location of the abnormality (Figure~\ref{fig:fractures1}). We also had access to descriptive sentences for each fracture retrieved from the original radiology reports, but these had highly inconsistent structure and content, making the task of training a system to generate similar sentences very difficult. 
To simplify the task, a radiologist created a new set of hand-labelled descriptive terms based on the image appearances. The elements of a radiology report we considered important in the diagnosis of hip fractures were the degree of fracture displacement, the degree of fragmentation (comminution), and the presence of subtle impaction (where one part of the bone is pushed into another), also shown in Figure~\ref{fig:fractures1}.
Unlike the task of identifying the presence or absence of fractures (a difficult task for humans~\cite{cannon2009imaging}), describing the visual features of a known fracture is trivial. For instance, the radiologist was able to label all of the 4,010 fractures with these descriptive terms (a total of 7 labels) in under 3 hours.

Because of the variation in language and quality in the original sentences, we use categorical variables for location and character to generate new sentences with a standard grammar and limited vocabulary. Each sentence had the general structure: \textbf{``There is a [degree of displacement], [+/- comminuted][+/- impacted] fracture of the [location] neck of femur [+/- with an avulsed fragment].''} Negative cases (i.e., those without fractures) were automatically labelled with a consistent sentence - "No fracture was identified on this study".
This sentence set was used to train an interpretability tool (described below), with the weights of the DenseNet held fixed. Therefore the image-analysis part of the model was never re-trained on the descriptive terms in the sentences (for example, to distinguish fractures with mild or severe displacement). We hypothesise that features relevant to this task were learned in the process of producing a fracture-detecting model.


\subsection{Model}
\label{sec:methods}

\begin{figure}
\begin{center}
\begin{tabular}{c}
\includegraphics[width=0.6\textwidth]{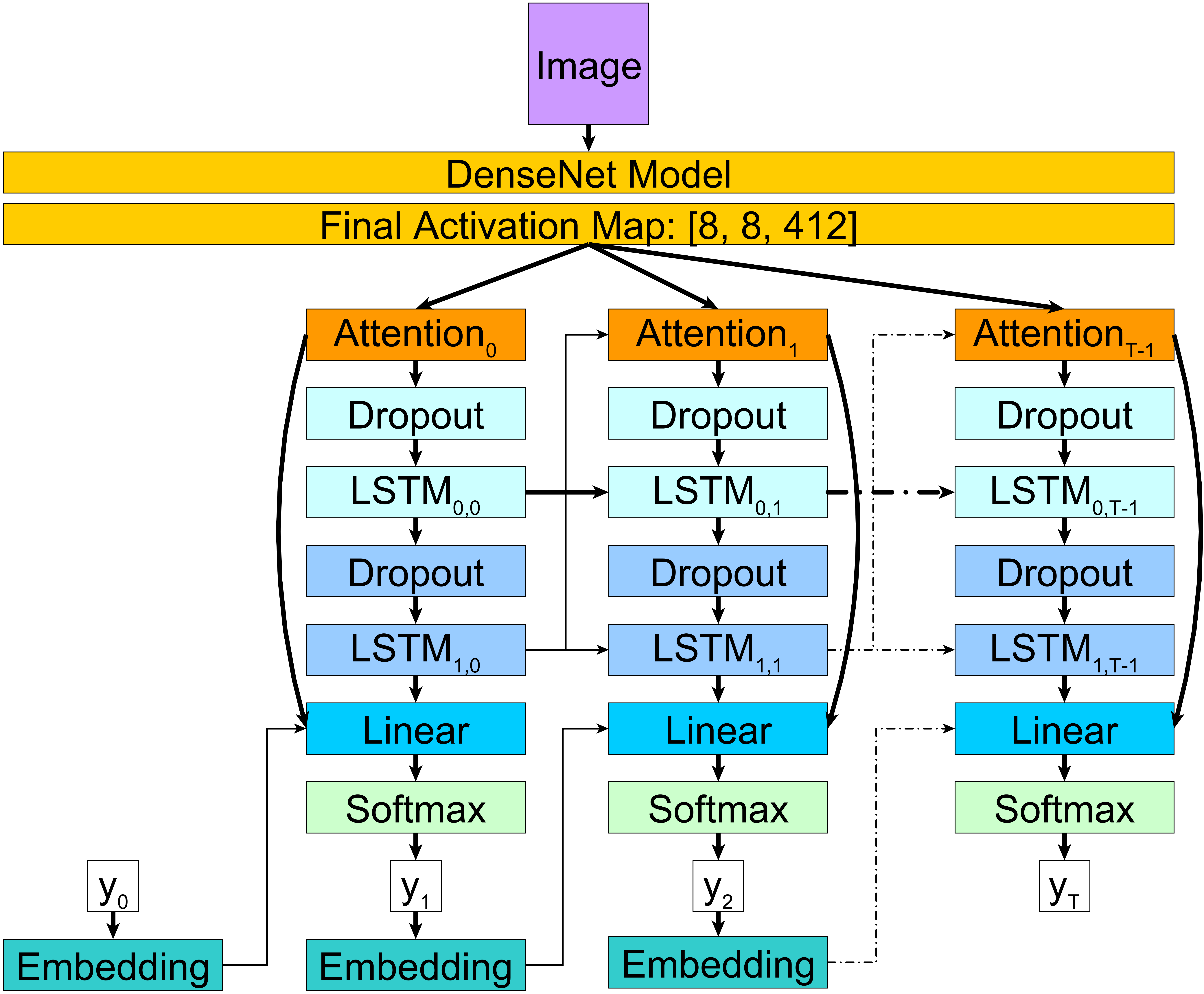} \\
\end{tabular}
\end{center}
\vspace{-.2in} 
\caption{The Architecture of the Image-to-Text Model}
\label{fig:captioning}
\end{figure}

Let us define the dataset as $\mathcal{D} = \{ (\mathbf{x}_i,\mathbf{y}_i,\mathbf{a}_i) \}_{i=1}^{|\mathcal{D}|}$, where $\mathbf{x}_i:\Omega \rightarrow \mathbb R$ denotes an X-ray image (with $\Omega$ representing the image lattice), 
$\mathbf{y}_i = [\mathbf{y}_i({1}),\dots,\mathbf{y}_i({T})]$, with $\mathbf{y}_i({t}) \in [0,1]^K$ - that is, $\mathbf{y}_i$ is a report containing $T$ words and the vocabulary size is $K$ (in this paper $K=30$, where the vocabulary comprises 26 lower-cased words plus start (SOT), end (EOT), unknown (UNK) and pad tokens), and
$\mathbf{a}_i = [\mathbf{a}_i(1),\dots,\mathbf{a}_i(C)]$, with $\mathbf{a}_i(c) \in \mathbb{R}^{D}$ ($C=8\times 8$ represents the number of image regions and $D=412$ denotes the region representation size), is the representation obtained from the trained DenseNet~\cite{gale2017detecting}, using the final activation map, i.e., the layer preceding average pooling and softmax.

Our proposed model agnostic interpretability method is defined by a recurrent neural network (RNN) model composed of two long short term memory (LSTM)~\cite{hochreiter1997long} layers, as depicted in Fig.\ref{fig:captioning}, where the LSTM layers are defined by
\begin{equation} 
\begin{split}
\label{eq:lstm}
\mathbf{h}_{0,t} &= LSTM_0(\mathbf{z}_{0,t},\mathbf{h}_{0,t-1}),\\
\mathbf{h}_{1,t} &= 
LSTM_1(\mathbf{z}_{1,t},\mathbf{h}_{1,t-1}), 
\end{split}
\end{equation}
where $\boldsymbol{\alpha}_{t} = [ \alpha_{t,1},\dots,\alpha_{t,C} ]$\ (with $\alpha_{t,c}\in [0,1]$) denotes a soft attention vector to be applied to each one of the $C$ regions of $\mathbf{a}$ and is estimated from inputs $\mathbf{h}_{1,t-1}$ and $\mathbf{a}$,
$\mathbf{z}_{0,t} = \text{dropout}(\boldsymbol{\alpha}_{t} \odot \mathbf{a})$ (we are abusing the notation for the elementwise multiplication $\odot$ by allowing each $\alpha_{t,c}$ to multiply the whole $D$-dimensional vector $\mathbf{a}(c)$, and dropout is defined in~\cite{srivastava2014dropout}), $\mathbf{z}_{1,t} = \text{dropout}(\mathbf{h}_{0,t})$
 and $\mathbf{h}_{0,t},\mathbf{h}_{1,t} \in \mathbb R^{512}$ are the hidden state representations. 
During inference, the initial hidden states are defined as $\mathbf{h}_{0,0}=\mathbf{h}_{1,0}=\vec{0}$, the initial word $\mathbf{y}({0})$ is a special SOT token and $\mathbf{y}({T})$ is the special EOT token.
The output layer~\cite{pascanu2013construct}, which predicts the next word in the sequence is defined by:
\begin{equation}
\widetilde{\mathbf{y}}_{t} = \text{softmax}(f_{\mathbf{W}}(\mathbf{h}_{1,t}, f_{\mathbf{E}}(\mathbf{y}_{t-1}), \boldsymbol{\alpha}_{t} \odot \mathbf{a})),
\end{equation}
where $\mathbf{W}$ and $\mathbf{E}$ are learnable parameters of the linear functions $f_{\mathbf{W}}(.)$ and $f_{\mathbf{E}}(.)$, respectively, with $f_{\mathbf{E}}(.)$ used to calculate a 256-dimensional word embedding. 
Both LSTM layers in (\ref{eq:lstm}) have learnable parameters~\cite{hochreiter1997long}, where the training process uses the length normalised cross-entropy loss, defined as follows for each training sample:
\begin{equation}
\ell = \frac{\sum_{t=1}^{L} \mathbf{y}_t^T\log(\widetilde{\mathbf{y}}_t)}{L}.
\end{equation}
The training is regularised using dropout with an 80\% keep-rate and data augmentation of the input image during training. 

\vspace{-.1in}
\section{Experiments and Results}
\label{sec:Experiments} 
\vspace{-.1in}

\subsection{BLEU Score}
\label{sec:BLEU score}
The BLEU score~\cite{papineni2002bleu} measures the similarity between the reference and model-generated sentences on the test set for 1-gram, 2-gram, 3-gram, and 4-gram. We calculated the BLEU score for the entire set of fractures in the held-out test set (348 images). To demonstrate the effect of simplifying our explanatory sentences, we show results for the same model architecture trained on either the original reports or our new sentences. We did not include the non-fracture test cases since the automatically generated "null" sentence would be flawlessly reproduced and identical for these samples, inflating the score.  The results (Table~\ref{tab:BLEU}), showed a high level of reproduction accuracy.  While these results are unsurprising (i.e., that an RNN can easily produce sentences with a simple structure and small vocabulary), they do confirm the hypothesis that these simplified sentences are highly learnable, despite the fact that CNN component was not trained on the descriptive terms in these sentences.

\begin{table}[h!]
\centering
\caption{BLEU scores on the fractures in the test set, comparing models trained to reproduce the original report sentences vs our simplified explanatory sentences.}
\label{tab:BLEU}
\begin{tabular}{ c @{\hskip 0.1in} c @{\hskip 0.1in} c}
\hline
& Original reports & Simplified sentences \\
\hline
1-gram & 65.0 & 91.9 \\  
\hline
2-gram & 37.9 & 83.8 \\  
\hline
3-gram & 24.2 & 76.1 \\  
\hline
4-gram & 15.9 &67.7 \\  
\hline
Weighted average & 25.67 & 77.97 \\  
\hline
\end{tabular}
\end{table}

\subsection{Sentence Content}
\label{sec:Sentence content}

The radiologist who created the original labels reviewed the sentences and images for 200 randomly chosen fractures from the test set in order to assess the semantic content of the sentences produced by our model. The radiologist reported the percentage of model-produced sentences and sentences from the original radiology reports that contained the appropriate location and character terms to describe the fractures. 

The text generation method was slightly worse at identifying the appropriate location, but better at describing the characteristics of the fracture (Table~\ref{tab:exp_set}).

\begin{table}[h!]
\centering
\caption{The percentage of explanatory sentences that contain an appropriate description of the location and character of the fractures, as determined by a radiologist.}
\label{tab:exp_set}
\begin{tabular}{ c c c}
\hline
&  Appropriate description &	Appropriate description  \\ 
&  of location &	of fracture character \\ 
\hline
Original radiology reports & 99\% & 78\% \\  
\hline
Our generated sentences & 90\% & 98\% \\  
\hline
\end{tabular}
\end{table}

The most common error in the generated sentences were ``off-by-one'' location errors, for example describing a subcapital fracture as transcervical. The most common error in the original radiology reports was failing to describe the character of the fracture.

\subsection{Acceptance of Explanations by Doctors}
\label{sec:Human assessment of explanations}
We presented several forms of explanations to 5 doctors, all with between 3 and 7 years of post-graduate clinical experience. We selected a subset of 30 cases from the test set, comprising 10 randomly selected fractures from the groups with mild, moderate, and severe displacement (to ensure a variety of examples). For each of these cases, we present 3 variations for explanations: 1) saliency maps produced by the SmoothGrad~\cite{smilkov2017smoothgrad} method, 2) sentences generated by our model, and 3) a combination of saliency maps and sentences. Each doctor was asked to assume they had received the diagnosis for each image from an unknown and untrusted source. They were asked to score each type of explanation on a Likert scale of 1 to 10, where 1 reflects a completely unsatisfactory explanation, and 10 reflects a perfect explanation. The average scores for each method are presented in Table~\ref{tab:doc_qual}, revealing that doctors prefer ``human-style'' text explanations over saliency maps, and have a preference for a combination of both saliency maps and generated text together compared to either alone. While the scale was arbitrary, the high scores consistently given to the combination of text and saliency maps suggest that the doctors were most satisfied with these type of explanations.

\begin{table}[h!]
\centering
\caption{The average qualitative score and the range of scores given by a group of 5 doctors for different explanations for a diagnosis of hip fracture.}
\label{tab:doc_qual}
\begin{tabular}{ c c}
\hline
&  Average qualitative score (range)  \\ 
\hline
Saliency maps alone & 4.4 (2-6) \\  
\hline
Generated sentences alone & 7.0 (6-8) \\  
\hline
Saliency maps and sentences & 8.8 (8-9) \\  
\hline
\end{tabular}
\end{table}

Finally, we present randomly selected examples of fractures, saliency maps, and both the original report descriptions and the generated text explanations (Figure ~\ref{fig:examples} ). Further examples are provided in the supplementary material.

\begin{figure}
\begin{center}
\begin{tabular}{c}
\includegraphics[width=0.75\textwidth]{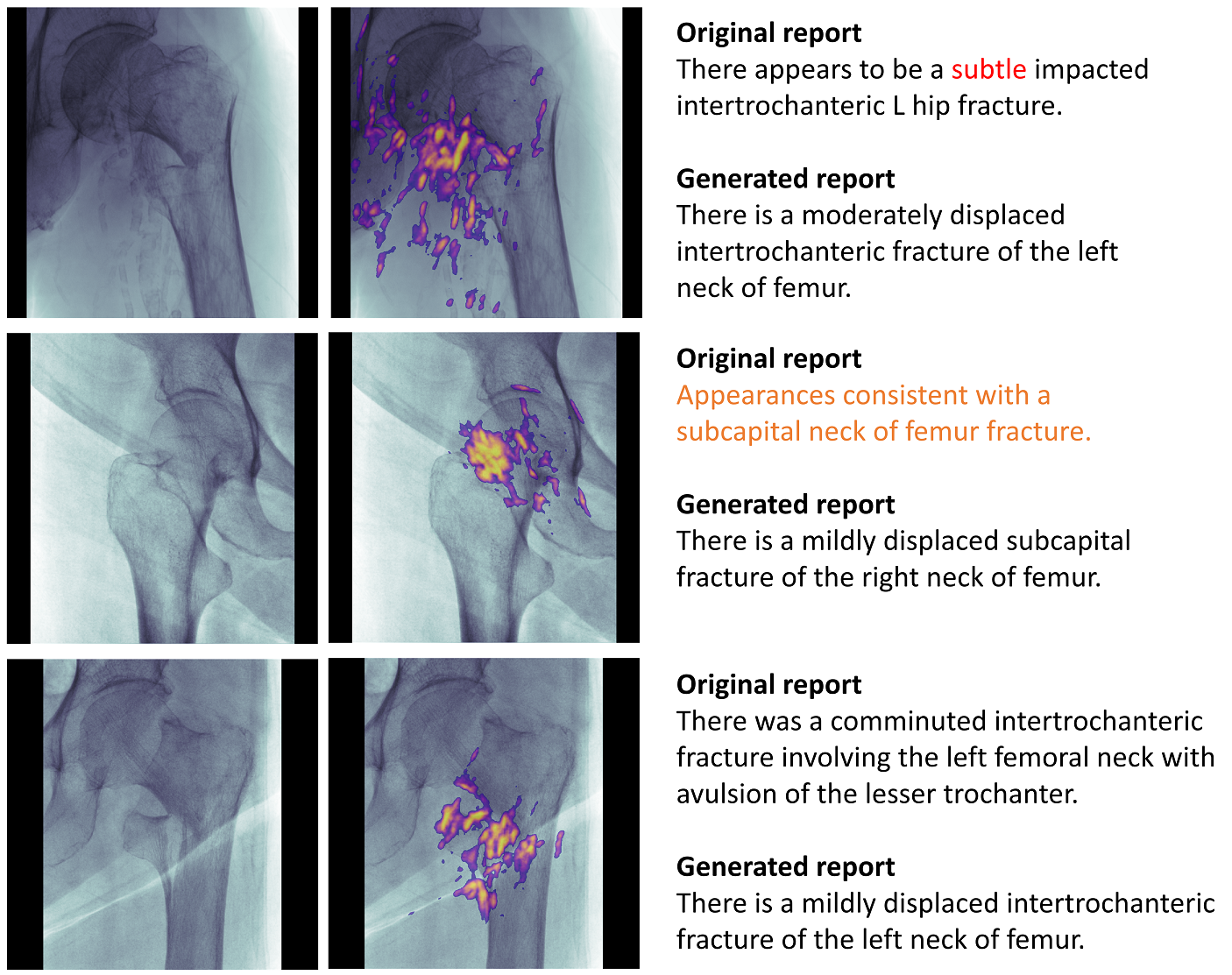} \\
\end{tabular}
\end{center}
\vspace{-.2in} 
\caption{Randomly selected examples, showing the original image, the SmoothGrad saliency map~\cite{smilkov2017smoothgrad}, the original report description, and the generated sentence. Orange highlights indicate that the sentence does not appropriately characterise the fracture. Red highlights indicate that the radiologist thought an alternative descriptive term was more appropriate.}
\label{fig:examples}
\end{figure}

\vspace{-.1in}
 \section{Discussion}
\label{sec:Discussion}
\vspace{-.1in}


We have shown experimentally that doctors prefer our generated text over saliency maps produced by a popular visualisation method, and prefer a combination of both over the generated sentences alone.

Our method is model-agnostic, meaning it can be applied to any deep learning image analysis model. We argue that a well-trained CNN that demonstrates human-level or greater performance at a medical task will already have learned to identify features to explain its decisions, and can be extended to produce a constrained text output for this task. Distilling explanatory sentences to their important elements makes the task tractable. 
The time cost of additional labelling appears to be manageable, at least in comparison to the effort of building the decision-making model. We believe this is because the descriptive terms are usually more easily identified and labelled than standard diagnostic tasks, and the labels required to build the diagnostic model will often overlap with the labels required to generate explanatory text. We expect this to be true in many medical settings; most individual image-based diagnoses in fields such as radiology, dermatology, and pathology are informed by very few key positive and negative findings, allowing for a similar approach to explanatory text generation.



\vspace{-.1in}
 \section{Conclusion}
\label{sec:Conclusion}
\vspace{-.1in}

Interpretable decision making is necessary in medicine, and machine learning systems that make medical decisions will be expected to address this need. 
Given that human-produced reports are considered sufficient as a form of explanation for medical decisions, we believe that our approach to generate text-based explanations provides a path to harmonise machine learning decisions and the human demand for explainability.

\end{document}